\documentclass[10pt, a4paper]{article}
\usepackage{lrec2022} % this is the new LREC2022 Style
\usepackage{multibib}
\newcites{languageresource}{Language Resources}
\usepackage{graphicx}
\usepackage{tabularx}
\usepackage{soul}
\usepackage{titlesec}
\titleformat{\section}{\normalfont\large\bfseries\center}{\thesection.}{1em}{}
\titleformat{\subsection}{\normalfont\SmallTitleFont\bfseries\raggedright}{\thesubsection.}{1em}{}
\titleformat{\subsubsection}{\normalfont\normalsize\bfseries\raggedright}{\thesubsubsection.}{1em}{}
\renewcommand\thesection{\arabic{section}}
\renewcommand\thesubsection{\thesection.\arabic{subsection}}
\renewcommand\thesubsubsection{\thesubsection.\arabic{subsubsection}}
\usepackage{epstopdf}
\usepackage[utf8]{inputenc}
\usepackage{hyperref}
\usepackage{xstring}
\usepackage{color}
\usepackage[T1]{fontenc}
\usepackage{microtype}

\title{Language Identification for Austronesian Languages}

\name{Jonathan Dunn, Wikke Nijhof} 

\address{University of Canterbury, Department of Linguistics \\
        and the New Zealand Institute for Language, Brain and Behaviour \\
         Christchurch, New Zealand \\
         \href{mailto:jonathan.dunn@canterbury.ac.nz}{jonathan.dunn@canterbury.ac.nz}, \href{mailto:wni20@uclive.ac.nz}{wni20@uclive.ac.nz} \\
         }

\abstract{
This paper provides language identification models for low- and under-resourced languages in the Pacific region with a focus on previously unavailable Austronesian languages. Accurate language identification is an important part of developing language resources. The approach taken in this paper combines 29 Austronesian languages with 171 non-Austronesian languages to create an evaluation set drawn from eight data sources. After evaluating six approaches to language identification, we find that a classifier based on skip-gram embeddings reaches a significantly higher performance than alternate methods. We then systematically increase the number of non-Austronesian languages in the model up to a total of 800 languages to evaluate whether an increased language inventory leads to less precise predictions for the Austronesian languages of interest. This evaluation finds that there is only a minimal impact on accuracy caused by increasing the inventory of non-Austronesian languages. Further experiments adapt these language identification models for code-switching detection, achieving high accuracy across all 29 languages. \\ \newline  }

\begin{document}

\maketitleabstract

\begin{table*}[t]
\centering
\begin{tabular}{|p{1.8cm}|p{1.2cm}|p{1.2cm}|p{1.2cm}|p{1.2cm}|p{1.2cm}|p{1.2cm}|p{1.8cm}|}
\hline
\textbf{Languages} & \textbf{langid.py} & \textbf{CLD3} & \textbf{fastText} & \textbf{polyglot} & \textbf{HeLI} & \textbf{idNet} & \textbf{This Paper} \\
\hline
Total & 97 & 107 & 176 & 196 & 285 & 464 & 200-800 \\
\hline
Austronesian & 5 & 7 & 9 & 12 & 20 & 10 & 29 \\
\hline
  \end{tabular}
  \caption{Comparison of Total and Austronesian Coverage for Common \textsc{lid} Models}
  \label{tab:1}
\end{table*}

\section{Identifying Pacific Languages}

Language identification (\textsc{lid}) remains an important problem within natural language processing because it is a central component in the creation of many corpora. The result is that languages which are not covered by a \textsc{lid} model also lack the data and resources necessary for many applications. This lack of data cannot be solved by bootstrapping methods \cite{bbfz09,geq12,b14} because it is not possible to identify new samples of the language. This means that accurate \textsc{lid} for minority languages has been and continues to be an important challenge \cite{Jauhiainen2019}, often the first step in developing resources for low-resource languages.

This paper addresses the problem of language identification for the Pacific region, with a special focus on low- and under-resourced Polynesian and Austronesian languages. We provide a \textsc{lid} model capable of accurately identifying 29 Polynesian/Austronesian languages against a selection of languages likely to occur in the Pacific region (200 in total). Of these Austronesian languages, 9 have not been previously available in \textsc{lid} models. An additional challenge is that the Austronesian languages of interest are closely related: similar, for example, to the challenge of identifying related Uralic languages \cite{chakravarthi-etal-2021-findings-vardial}. This makes the identification task more difficult because the languages themselves are relatively similar. Even for Austronesian languages that have been included in previous models, then, this work improves our ability to distinguish them from a wider range of closely related languages.

The experiments in this paper first evaluate six \textsc{lid} architectures on an inventory of 200 languages to see which approaches work best with a focus on Austronesian languages. These experiments show that a classifier based on skip-gram character embeddings performs best by a clear margin. We then evaluate this approach against an increasing inventory of languages, from 200 to 800, in order to quantify the trade-off between accuracy and coverage \cite{Majlis2012,jauhiainen-etal-2017-evaluating}. 

The problem of language identification is often presented as a trade-off between (i) the number of languages, (ii) the sample size for each document, and (iii) the diversity of data sources included. For example, it is often possible to maintain high accuracy for a very large number of languages if the training and testing data is drawn from a limited set of domains \cite{Brown2014}. This paper uses data from many different domains and conducts an evaluation on a test set containing 1.1 million samples. The combination of many domains and many test samples helps to ensure that the evaluation represents a realistic context: for example, the application of a \textsc{lid} model in a bootstrapping context \cite{Dunn2020} requires accurate identification over a very large number of samples that usually show a highly skewed distribution of labels.

This paper maintains high accuracy at a sample size of 100 characters per document while continuing to cover a large number of Austronesian languages. Previous work has evaluated \textsc{lid} models on sample sizes as small as 40 or 60 characters \cite{Brown2014,jauhiainen-etal-2017-evaluation}. A smaller sample size like this requires a trade-off, either in reducing the inventory of languages or reducing the diversity of domains used for training and testing. The longer-term goal for this work is to enable bootstrapping methods to develop corpora representing the Pacific region. Such bootstrapped corpora collect data from many domains; thus, our focus is on maintaining a diversity of domains and a large language inventory rather than reducing sample size. Further, documents with fewer than 100 characters are often less useful from a corpus-building perspective.

We begin, in Section 2, by reviewing the selection of non-Austronesian languages to include in our initial inventory. We then describe our sources of data (Section 3) and the specific models as they have been implemented here (Section 4). Further, we discuss the evaluation across different models in Section 5 and the evaluation across different language inventories in Section 6. In Section 7 we apply these language identification models to code-switching detection and in Section 8 we evaluate the performance and stability of models after compression.

\section{Inventory of Languages}

One challenge for the identification of low-resource languages is that there are so many such languages that the inventory can become quite large relative to the amount of training data available. This paper first evaluates different methods on an inventory of only 200 languages, including 171 non-Austronesian languages. This section details previous coverage of the relevant languages as well as how we select the non-Austronesian languages to include in the initial model.

We have two types of constraints: First, any Austronesian language, our main focus, is included in the language inventory by default. This is based on genetic classifications. Second, recent work on digital language mapping \cite{Dunn2020,dunn-adams-2020-geographically} has used web and social media data to determine the inventory of languages most likely to occur in each region, including the Pacific. We take the non-Austronesian languages that are most common in the Pacific region in this previous work\footnote{https://www.earthLings.io} and include them in the initial model. This provides an inventory of 200 languages; 29 of these are Austronesian and the remainder are those frequently observed in the Pacific. A complete list is available in the supplementary material.

In Table \ref{tab:1} this inventory is compared with six common \textsc{lid} packages: Google's CLD3\footnote{https://github.com/google/cld3}, langid.py\footnote{https://github.com/saffsd/langid.py}, polyglot\footnote{https://github.com/aboSamoor/polyglot}, fastText\footnote{https://fasttext.cc/docs/en/language-identification.html},
idNet\footnote{https://github.com/jonathandunn/idNet}, and HeLI\footnote{https://github.com/tosaja/HeLI}. Most work that depends on language identification indirectly relies on one or another of these packages. The table shows the total inventory of languages for each model as well as the number of Austronesian languages. The base model evaluated here includes 200 languages in total, more than any package other than idNet and HeLI. The largest inventory, 800 languages, includes more than any of the other packages.

\section{Data}

The ground-truth corpora used as samples for each language are taken from several sources, detailed in Table \ref{tab:2}. Corpora are split into samples of 100 characters and cleaned using the \textit{clean-text} package\footnote{https://pypi.org/project/clean-text/} to remove \textsc{urls}, numbers, punctuation, and other non-linguistic characters. We divide the data into training, testing, and validation sets. Within each family of models, discussed in Section 4, we first find the best parameters using the test set and then conduct the evaluation on the validation set. These data sources provide a diversity of domains that is important for ensuring a robust evaluation of \textsc{lid} performance. This also provides a very large training set (over 100 million samples) and a very large testing set (over 1.1 million samples).

\begin{table}[h]
\centering
\begin{tabular}{|l|r|}
\hline
\textbf{Corpus} & \textbf{N. Langs} \\
\hline
Bible Translations & 614 \\
\cite{Brown2014} & ~ \\
\hline
Global Voices News & 41 \\
\cite{Tiedemann2012} & ~  \\
\hline
JW 300 & 380 \\
\cite{agic-vulic-2019-jw300} & ~ \\
\hline
Open Subtitles & 62 \\
\cite{lison-tiedemann-2016-opensubtitles2016} & ~ \\
\hline
QCRI Educational Domain & 42 \\
\cite{Tiedemann2012} & ~ \\
\hline
Tatoeba Sentences & 309 \\
\cite{Tiedemann2012} & ~ \\
\hline
Wikipedia Articles & 280 \\
TensorFlow DataSets & ~ \\
\hline
M\={a}ori Broadcasts & 1 \\
\cite{Boyce2006} & ~ \\
\hline
  \end{tabular}
  \caption{Sources of Data for \textsc{lid} Models}
  \label{tab:2}
\end{table}

\section{Models}

\begin{table*}[t]
\centering
\begin{tabular}{|p{1.6cm}|p{.9cm}p{.9cm}|p{.9cm}p{.9cm}|p{.9cm}p{.9cm}|p{.9cm}p{.9cm}|p{.9cm}p{.9cm}|}
\hline
~ & \multicolumn{2}{|c|}{\textbf{NB, InfoGain}} & \multicolumn{2}{|c|}{\textbf{SVM, InfoGain}} & \multicolumn{2}{|c|}{\textbf{MLP, InfoGain}} & \multicolumn{2}{|c|}{\textbf{MLP, Hashing}} & \multicolumn{2}{|c|}{\textbf{fastText}} \\
\hline
\textbf{Language} & \textbf{Prec.} & \textbf{Rec.} & \textbf{Prec.} & \textbf{Rec.} & \textbf{Prec.} & \textbf{Rec.} & \textbf{Prec.} & \textbf{Rec.} & \textbf{Prec.} & \textbf{Rec.} \\
\hline
\textit{W. Average} & \textit{0.95} & \textit{0.94} & \textit{0.96} & \textit{0.95} & \textit{0.95} & \textit{0.93} & \textit{0.96} & \textit{0.95} & \textit{0.99} & \textit{0.99} \\
\hline
Acehnese & 0.99 & 0.98 & 1.00 & 0.90 & 1.00 & 0.84 & 0.99 & 0.96 & 1.00 & 0.99 \\
Buginese & 1.00 & 0.94 & 1.00 & 0.92 & 1.00 & 0.93 & 1.00 & 0.94 & 1.00 & 0.98 \\
Cebuano & 0.95 & 0.88 & 0.98 & 0.99 & 0.69 & 0.99 & 0.69 & 0.89 & 1.00 & 1.00 \\
Chamorro & 1.00 & 0.64 & 1.00 & 0.31 & 0.99 & 0.91 & 1.00 & 0.93 & 0.96 & 1.00 \\
Chuukese & 1.00 & 1.00 & 1.00 & 1.00 & 0.99 & 1.00 & 0.99 & 1.00 & 1.00 & 1.00 \\
Fijian & 1.00 & 0.92 & 1.00 & 0.93 & 1.00 & 0.96 & 1.00 & 0.94 & 1.00 & 1.00 \\
Gilbertese & 1.00 & 1.00 & 1.00 & 1.00 & 1.00 & 0.98 & 1.00 & 0.97 & 1.00 & 1.00 \\
Hawaiian & 1.00 & 0.99 & 1.00 & 0.97 & 0.99 & 0.99 & 0.97 & 1.00 & 1.00 & 1.00 \\
Hiligaynon & 0.64 & 0.98 & 0.97 & 0.98 & 0.05 & 0.00 & 0.73 & 0.01 & 0.99 & 1.00 \\
Hiri Motu & 1.00 & 1.00 & 1.00 & 1.00 & 0.98 & 1.00 & 0.99 & 1.00 & 1.00 & 1.00 \\
Ilocano & 0.97 & 0.97 & 0.99 & 0.98 & 0.99 & 0.97 & 0.99 & 0.97 & 1.00 & 0.99 \\
Javanese & 0.66 & 0.98 & 0.83 & 0.97 & 0.99 & 0.87 & 1.00 & 0.74 & 0.97 & 0.99 \\
Marshallese & 1.00 & 1.00 & 1.00 & 1.00 & 1.00 & 0.99 & 1.00 & 1.00 & 1.00 & 1.00 \\
Malagasy & 0.99 & 0.99 & 1.00 & 1.00 & 0.98 & 1.00 & 0.95 & 1.00 & 1.00 & 1.00 \\
Māori & 0.92 & 0.99 & 1.00 & 0.98 & 0.74 & 1.00 & 0.68 & 0.99 & 1.00 & 1.00 \\
Malay & 0.97 & 0.98 & 0.95 & 0.99 & 0.91 & 0.98 & 0.84 & 0.97 & 0.97 & 0.99 \\
Niuean & 1.00 & 1.00 & 1.00 & 1.00 & 0.94 & 1.00 & 0.97 & 0.96 & 1.00 & 1.00 \\
Pangasinan & 0.98 & 0.86 & 0.98 & 0.94 & 0.99 & 0.93 & 0.98 & 0.89 & 1.00 & 0.97 \\
Pohnpeian & 1.00 & 1.00 & 1.00 & 1.00 & 1.00 & 1.00 & 1.00 & 1.00 & 1.00 & 1.00 \\
C.I. Māori & 0.99 & 1.00 & 0.99 & 0.99 & 0.55 & 0.01 & 0.38 & 0.00 & 1.00 & 1.00 \\
Samoan & 1.00 & 0.97 & 1.00 & 0.98 & 0.96 & 0.98 & 0.99 & 0.96 & 1.00 & 0.99 \\
Sundanese & 0.97 & 0.93 & 0.95 & 0.93 & 0.88 & 0.91 & 0.94 & 0.80 & 1.00 & 0.96 \\
Tahitian & 0.99 & 0.72 & 0.99 & 0.84 & 1.00 & 0.94 & 1.00 & 0.81 & 1.00 & 1.00 \\
Tagalog & 0.94 & 0.98 & 0.95 & 0.98 & 0.98 & 0.92 & 0.89 & 0.97 & 0.99 & 0.99 \\
Tongan & 1.00 & 0.92 & 1.00 & 0.92 & 0.86 & 0.93 & 0.90 & 0.93 & 1.00 & 0.97 \\
Tuvaluan & 1.00 & 1.00 & 1.00 & 1.00 & 1.00 & 0.87 & 1.00 & 0.86 & 1.00 & 1.00 \\
Waray & 0.97 & 0.86 & 1.00 & 1.00 & 1.00 & 0.94 & 0.99 & 0.89 & 1.00 & 1.00 \\
Wallisian & 1.00 & 1.00 & 1.00 & 0.99 & 0.97 & 0.56 & 0.95 & 0.61 & 1.00 & 1.00 \\
Yapese & 1.00 & 1.00 & 1.00 & 1.00 & 1.00 & 0.99 & 0.99 & 1.00 & 1.00 & 1.00 \\
\hline
  \end{tabular}
  \caption{Break-Down of \textsc{lid} Performance by Model and Language (200 Language Inventory)}
  \label{tab3}
\end{table*}

Based on previous work on language identification, we implement six main approaches, a combination of neural and non-neural architectures. While many popular packages rely on neural models (such as Google's CLD3), non-neural models often dominate shared tasks \cite{chakravarthi-etal-2021-findings-vardial}. Most work on \textsc{lid} uses character n-grams as features, usually trigrams or ranges of \textit{n} that include trigrams. Some work has then focused on the problem of feature selection across all possible character n-grams to produce a useful feature space \cite{lui-baldwin-2011-cross}. Here we implement a similar information gain feature selection method, experimenting with the number of features and the n-gram range on the test set. That same feature set is then used across all relevant classifiers. Here the best variant uses information gain to choose the top 75k character trigrams.

First, we implement a feed-forward network that is similar to both CLD3 and idNet \cite{Dunn2020}. The specific architecture of this network is determined on the test set, ultimately containing two layers of 500 ReLU neurons together with a softmax prediction layer. This is listed as \textit{MLP, InfoGain} in Table \ref{tab3}. The specific implementations of this first model and the second model (below) are provided in the supplementary material.

Second, recent work has shown that feature hashing for character n-grams works well for language identification \cite{Malmasi2017,Dunn2020}, sometimes better than class-based feature selection. For example, two of the comparison systems in Table \ref{tab:1} use feature hashes within a feed-forward network with softmax for predictions (CLD3, idNet). Here the best hash-based classifier uses a hashing space with 150k bins within a feed-forward network with one layer of 200 neurons and a softmax prediction layer. This is listed as \textit{MLP, Hashing} in Table \ref{tab3}. The substantial difference between these first and second variants, then, is whether the character n-grams are derived from feature selection or from a hashing algorithm; the code for both is available in the supplementary material.

\begin{table*}[t]
\centering
\begin{tabular}{|p{1.5cm}|p{1.6cm}p{2.1cm}|p{1.6cm}p{2.1cm}|p{1.6cm}p{2.1cm}|}
\hline
\textbf{Languages} & \textbf{Prec. (All)} & \textbf{Prec. (Pacific)} & \textbf{Rec. (All)} & \textbf{Rec. (Pacific)} & \textbf{F1 (All)} & \textbf{F1 (Pacific)} \\
\hline
200 & 0.994 & 0.995 & 0.994 & 0.993 & 0.994 & 0.994 \\
300 & 0.973 & 0.956 & 0.973 & 0.994 & 0.969 & 0.974 \\
400 & 0.970 & 0.963 & 0.970 & 0.988 & 0.965 & 0.973 \\
500 & 0.969 & 0.967 & 0.970 & 0.989 & 0.965 & 0.977 \\
600 & 0.971 & 0.979 & 0.963 & 0.935 & 0.960 & 0.946 \\
700 & 0.969 & 0.964 & 0.968 & 0.993 & 0.964 & 0.976 \\
800 & 0.966 & 0.970 & 0.968 & 0.979 & 0.962 & 0.974 \\
\hline
  \end{tabular}
  \caption{Decreasing Performance With Increasing Language Inventory}
  \label{tab:4}
\end{table*}

Third, an approach based on skip-gram character embeddings, as implemented in the fastText package, has been shown to be effective for language identification \cite{joulin2016bag}. Here the best model is based on character skip-gram embeddings, with n-grams ranging from 1 to 4 with 300 dimensions and 100 negative samples. This is listed as \textit{fastText} in Table \ref{tab3}.

Fourth, recent results from the VarDial evaluation campaign focus on a related problem for Uralic languages, in which 29 region-specific languages of interest are combined with 149 non-relevant languages for the task of distinguishing similar languages \cite{chakravarthi-etal-2021-findings-vardial}. While Uralic languages are quite different from Austronesian languages, the underlying problem remains comparable. In this evaluation campaign, many of the best approaches used non-neural classifiers such as Naive Bayes or Support Vector Machines together with character n-grams. We evaluate both of these approaches in Table \ref{tab3}, each using the same feature set as the first feed-forward network. While these classifiers have worked well on problems with limited training data, the training set here contains over 100 million samples. SVMs, in particular, are difficult to train in this setting. Thus, we create a more practical sub-set of the training set, containing 50 million samples (for Naive Bayes) and 1 million samples (for the SVM). These variants are listed as \textit{NB} and \textit{SVM} in Table \ref{tab3}.

Fifth, sequence-based models have used LSTM networks to make predictions directly on sequences of characters, avoiding the need for selecting character n-grams as features \cite{jaech-etal-2016-hierarchical,kocmi-bojar-2017-lanidenn}. These models have typically worked with a much smaller inventory of languages, often with a focus on code-switching. Here we have experimented with several variants of a sequence-based LSTM network. However, none of these variants achieved a competitive accuracy. Thus, we focus our evaluation on the five methods above, all based on character n-grams: a feed-forward network, Naive Bayes, and a Support Vector Machine, all with feature selection; a feed-forward network with feature hashing; and a classifier based on skip-gram character embeddings.

\section{Evaluation}

We first evaluate each of these models using precision and recall, as shown in Table \ref{tab3}. The first row shows the weighted average across all 200 languages. The data is drawn from multiple sources, so that high-resource languages which appear in each source tend to contribute more test samples. The total validation set contains over 1.1 million samples, with each language ranging from 2k to 16k samples. Given the weighted precision and recall, the \textsc{sgns}-based classifier represented by fastText out-performs the other models, reaching 0.99 for both precision and recall.

The table also shows each of the Austronesian languages separately in order to determine whether this average performance is representative of these low-resource languages. For example, the feature selection MLP model performs poorly for several languages: Cook Islands M\={a}ori and Hiligaynon most prominently. And the feature hashing MLP model produces poor results for Javanese and Wallisian. The \textsc{sgns} model, however, never falls below 0.96 for any Austronesian language. Thus, not only does the model provide overall accurate predictions at a small sample size, but that performance remains robust across the low-resource languages we are concerned with. In this setting, neither the Naive Bayes nor the SVM models exhibit  the best performance.

\section{Influence of Inventory Size}

\begin{table*}[t]
\centering
\begin{tabular}{|p{1.2cm}p{1.0cm}p{1.0cm}p{1.0cm}p{1.0cm}p{1.0cm}p{1.0cm}p{1.0cm}p{1.0cm}p{1.0cm}|}
\hline
awesome & video & diaries & ka & mau & te & wehi & e & te & whanau \\
\textsc{eng} & \textsc{eng} & \textsc{eng} & \textsc{mri} & \textsc{mri} & \textsc{mri} & \textsc{mri} & \textsc{mri} & \textsc{mri} & \textsc{mri} \\
\hline
  \end{tabular}
  \caption{Code-Switching, English and te reo M\={a}ori}
  \label{tab:6}
\end{table*}

How well would this approach have worked if we had instead chosen an inventory of 300 or 500 or 800 languages? We evaluate this in Table \ref{tab:4}; each model contains the same languages as previous models together with a selection of additional languages. Languages are added in order of the number of samples available in the training data, so that the later languages are those with the fewest available samples. The models with 200 and 800 languages are made available for further use.\footnote{https://www.jdunn.name/corpora/}

The final column in the table shows the weighted f-score, which decreases from 0.994 to 0.962 (all languages) and 0.974 (Austronesian languages) as the number of languages increases. This would seem to show a relatively minor reduction in performance given increased coverage. For example, this performance remains higher than the other methods evaluated in Section 5. While precision and recall show a similar slight reduction, there is variation in the precision for Austronesian languages: the metric both increases and decreases. This indicates that the performance of the \textsc{sgns} architecture is somewhat unstable, perhaps caused by variability in this type of embedding \cite{burdick-etal-2021-analyzing}. The fact that the smaller set of Austronesian languages is more subject to variation indicates that variation in the main inventory of languages is averaged out. This variability in performance is investigated further in Section 8; it remains the case that the best-performing model for identifying Austronesian languages is the \textsc{sgns} approach.

\begin{table*}[h]
\centering
\begin{tabular}{|p{5.0cm}|p{2.5cm}|p{2.5cm}|p{2.5cm}|}
\hline
\textbf{~} & \textbf{MLP, Selection} & \textbf{fastText} & \textbf{fastText (Small)} \\
%Austronesian & Acehnese & 99.83\% & 98.20\% & 99.05\% \\
%Austronesian & Cebuano & 96.33\% & 99.70\% & 99.60\% \\
%Austronesian & Fijian & 98.37\% & 98.40\% & 99.85\% \\
%Austronesian & Gilbertese & 99.74\% & 99.90\% & 99.90\% \\
%Austronesian & Hiligaynon & 97.73\% & 99.90\% & 99.78\% \\
%Austronesian & Hiri & 99.18\% & 99.90\% & 99.83\% \\
%Austronesian & Ilocano & 97.43\% & 99.30\% & 96.16\% \\
%Austronesian & Javanese & 93.06\% & 99.10\% & 99.00\% \\
%Austronesian & Malagasy & 97.51\% & 98.80\% & 99.42\% \\
%Austronesian & Malay & 95.13\% & 99.00\% & 98.93\% \\
%Austronesian & Sundanese & 89.25\% & 97.90\% & 98.54\% \\
%Austronesian & Tagalog & 98.15\% & 98.90\% & 98.46\% \\
%\hline
%Polynesian & Cook Islands M\={a}ori & 99.91\% & 99.90\% & 99.90\% \\
%Polynesian & Hawaiian & 99.81\% & 99.60\% & 99.25\% \\
%Polynesian & M\={a}ori & 99.79\% & 98.40\% & 99.34\% \\
%Polynesian & Samoan & 99.68\% & 99.30\% & 99.04\% \\
%Polynesian & Tahitian & 99.70\% & 98.00\% & 99.83\% \\
%Polynesian & Tongan & 97.95\% & 98.30\% & 97.63\% \\
%Polynesian & Tuvaluan & 99.88\% & 99.90\% & 99.93\% \\
\hline
{Average Across 19 Languages} & {97.81\%} & {99.07\%} & {99.13\%} \\
\hline
  \end{tabular}
  \caption{Accuracy of Short-Span Detection by Model (20 characters)}
  \label{tab:5}
\end{table*}

This section has evaluated models for identifying a set of closely related and previously unavailable Austronesian languages. The evaluation has shown that, while the \textsc{sgns} model is subject to some variation, it remains the best performing approach against other models, with an f-score of 0.974 when evaluated with a total inventory of 800 languages.

\section{Code-Switching Detection}

An additional problem closely related to language identification is the detection of code-switching within samples \cite{solorio-etal-2014-overview,molina-etal-2016-overview}. For example, Table \ref{tab:6} shows a tweet which is roughly one-third English (\textsc{eng}) and two-thirds te reo M\={a}ori (\textsc{mri}). This type of code-switching is especially common in digital contexts like social media, the precise contexts from which most training sets are derived. Corpora in te reo M\={a}ori often contain a significant amount of English words, as do corpora in many other Austronesian languages. For language identification to be useful for the purpose of building corpora, it is important to be able to also identify code-switching within a corpus. This section presents experiments in code-switching detection for Austronesian languages based on the language identification models presented in Section 4.

Given the previous approach to language identification at 100-character spans, the basic approach taken to code-switching detection is based on disaggregation: \textit{First}, we train language identification models for English and each of the 29 languages in Table \ref{tab:5a}; these models include only two languages, but have a much shorter character span of 15-characters. We evaluate the performance of short-span detection by model type in Table \ref{tab:5} and then by language for the best model in Table \ref{tab:5a}. \textit{Second}, we use predictions on overlapping spans to convert this span-based prediction to a word-by-word prediction. This section first evaluates different models for short-span language identification before presenting and evaluating two algorithms for converting short-span identification into word-by-word code-switching detection for Austronesian languages.

\subsection{Evaluating Short-Span Identification}

\begin{table}[h]
\centering
\begin{tabular}{|lccr|}
\hline
\textbf{Language} & \textbf{Prec.} & \textbf{Rec.} & \textbf{Support} \\
\hline
Acehnese & 0.99 & 0.98 & 11,005 \\
Buginese & 0.98 & 0.99 & 10,547 \\
Cebuano & 1.00 & 1.00 & 33,112 \\
Chamorro & 1.00 & 0.99 & 11,061 \\
Chuukese & 1.00 & 1.00 & 22,518 \\
Fijian & 1.00 & 1.00 & 11,185 \\
Gilbertese & 1.00 & 1.00 & 11,225 \\
Hawaiian & 1.00 & 0.99 & 22,557 \\
Hiligaynon & 1.00 & 0.99 & 11,133 \\
Hiri Motu & 1.00 & 1.00 & 22,593 \\
Ilocano & 0.99 & 0.98 & 21,620 \\
Javanese & 0.99 & 0.98 & 22,094 \\
Marshallese & 1.00 & 1.00 & 11,403 \\
Malagasy & 0.99 & 0.99 & 32,220 \\
Māori & 1.00 & 0.99 & 34,023 \\
Malay & 0.99 & 0.98 & 43,967 \\
Niuean & 1.00 & 1.00 & 11,325 \\
Pangasinan & 1.00 & 0.95 & 22,011 \\
Pohnpeian & 1.00 & 1.00 & 22,182 \\
C.I. Māori & 1.00 & 1.00 & 11,425 \\
Samoan & 1.00 & 0.98 & 34,053 \\
Sundanese & 0.98 & 0.96 & 21,967 \\
Tahitian & 1.00 & 1.00 & 11,537 \\
Tagalog & 0.99 & 0.98 & 55,022 \\
Tongan & 1.00 & 0.94 & 33,790 \\
Tuvaluan & 1.00 & 1.00 & 11,468 \\
Waray & 0.99 & 1.00 & 21,422 \\
Wallisian & 1.00 & 1.00 &  11,455 \\
Yapese & 1.00 & 1.00 & 23,034 \\
\hline
  \end{tabular}
  \caption{Performance of Short-Span Detection by Language with fastText (compressed models, 15 characters)}
  \label{tab:5a}
\end{table}

We begin by evaluating the accuracy for the top two models for language identification when applied to the short-span identification task. Here the short-span task involves distinguishing between two languages (English and an Austronesian language) with a window of 15 characters. The idea behind this algorithm is to support corpus cleaning: first, we identify that a particular document belongs to an Austronesian language; second, we search for English material within that document.

\begin{table*}[t]
\centering
\begin{tabular}{|l|l|}
\hline
~ & \textbf{Variables}\\
\hline
~ & \textit{sample} = Current document or sentence \\
~ & \textit{word} = Unit separated by white space after tokenization \\
~ & \textit{character} = Individual symbol within word \\
~ & \textit{character span} = Window of 15 characters that ignores word boundaries \\
~ & \textit{trigram context} = Current word along with one previous word and one following word \\
\hline
\hline
~ & \textbf{Algorithm 1: Overlapping Character Spans} \\
\hline
1 & For each word in sample: \\
2 & \hspace{0.5cm} For each character in word: \\
3 & \hspace{1cm} $p(lang)$ = Probability of English predicted for current character span \\
4 & \hspace{0.5cm} Return Mean($p(lang)$) = Average probability of English across entire word\\
\hline
\hline
~ & \textbf{Algorithm 2: Word-Based Detection} \\
\hline
1 & $p(span)$ = Probability of English for current sample\\
2 & For each word in sample: \\
3 & \hspace{0.5cm} $p(word)$ = Probability of English for current word alone\\
4 & \hspace{0.5cm} $p(trigram)$ Probability of English for trigram context around current word\\
5 & \hspace{0.5cm} If $p(span)$ closer to English:\\
6 & \hspace{1.0cm} Return max($p(word)$, $p(trigam)$)\\
7 & \hspace{0.5cm} Else If $p(span)$ further from English:\\
8 & \hspace{1.0cm} Return min($p(word)$, $p(trigam)$)\\
\hline
  \end{tabular}
  \caption{Code-Switching Detection Algorithms}
  \label{tab:7}
\end{table*}

The results for this short-span identification task are shown in Table \ref{tab:5}; given the performance reported above, we focus on fastText and the MLP with feature selection methods as the most promising. As before, the average performance for fastText is slightly higher across all Austronesian languages, with 99\% vs 97.8\%. There are several instances where fastText performs significantly higher than the MLP: Javanese (+6.04\%), Malay (+3.87\%), and Sundanese (+8.65\%). We also include a compressed fastText model for comparison to ensure that model size does not impact prediction accuracy. 

After comparing the performance on the short-span task, then, we proceed with code-switching detection using the fastText models. We show the evaluation of compressed fastText models in Table \ref{tab:5a} at a span-size of 15 characters. This table shows the precision and recall specifically for the Austronesian languages as well as the number of test samples for each language. The performance here is consistently high, with only two languages falling below 0.99 precision (Buginese and Sundanese). Tongan is the only language with a recall below 0.95, so that some samples from Tongan are identified as English instead.

\subsection{Algorithms for Word-Level Prediction}

The short-span prediction task evaluated above allows us to perform language identification on very small samples, but it does not directly provide code-switching detection at the word-level. We therefore evaluate two approaches to converting short-span predictions into word-level predictions, as shown in Table \ref{tab:7}.

The first algorithm is based on averaging predictions across overlapping character spans. \textit{First}, the algorithm tokenizes the sample into words and iterates over words; this is required to make predictions about the language for each word in the sample. \textit{Second}, for each word, the algorithm predicts the probability that each span centered on a character within that word belongs to either English or the Austronesian language, using the short-span identification model. Thus, a word containing five characters is represented by five character spans. The algorithm then returns the average probability across all spans as the word's overall value.

The examples in (a) through (d) show a selection of the character spans that are queried for each word. Cases like \textit{awesome} in (a) and \textit{mau} in (d) are straight-forward in the sense that only words from one language fall within the 15-character span. The challenge for this algorithm comes from examples like (b) and (c), in which the character spans around the word contain code-switching within them. A baseline algorithm in the evaluation simply queries each word in the short-span model directly. The intuition behind this first span-based algorithm is that words common in both languages (for example, \textit{i} and \textit{he} in the case of \textsc{eng-mri} code-switching) will be identified using information about their immediate context as well.

~

\hspace{1cm}(a) \textit{awesome} = awesome\_video\_diarie
~

\hspace{1cm}(b) \textit{diaries} = video\_diaries\_ka\_ma\_
~

\hspace{1cm}(c) \textit{ka} =  \_diaries\_ka\_mau\_te\_w
~

\hspace{1cm}(d) \textit{mau} = aries\_ka\_mau\_te\_wehi

~

\begin{table}[h]
\centering
\begin{tabular}{|l|c|c|c|}
\hline
\textbf{Language} & \textbf{Alg. 1} & \textbf{Alg. 2} & \textbf{Baseline} \\
\hline
Cebuano & 24\% & 100\% & 94\% \\
Chamorro & 96\% & 100\% & 100\% \\
Fijian & 54\% & 98\% & 48\% \\
Hawaiian & 64\% & 100\% & 94\% \\
Ilocano & 74\% & 100\% & 100\% \\
Javanese & 92\% & 100\% & 100\% \\
Malagasy & 50\% & 100\% & 100\% \\
Māori & 40\% & 100\% & 60\% \\
Malay & 96\% & 100\% & 100\% \\
Samoan & 34\% & 100\% & 44\% \\
Sundanese & 92\% & 100\% & 100\% \\
Tahitian & 56\% & 98\% & 86\% \\
Tagalog & 94\% & 100\% & 100\% \\
Tongan & 70\% & 100\% & 68\% \\
\hline
\textbf{Average} & \textbf{67\%} & \textbf{100\%} & \textbf{85\%} \\
\hline
  \end{tabular}
  \caption{Accuracy of Identified English Spans in Wikipedia}
  \label{tab:8}
\end{table}

\begin{table*}[t]
\centering
\begin{tabular}{|p{1.6cm}|p{.9cm}p{.9cm}|p{.9cm}p{.9cm}|p{.9cm}p{.9cm}|p{.9cm}p{.9cm}|p{.9cm}p{.9cm}|}
\hline
~ & \multicolumn{2}{|c|}{\textbf{Original, 200L}} & \multicolumn{2}{|c|}{\textbf{100d, Ftz, 200L}} & \multicolumn{2}{|c|}{\textbf{200d, Ftz, 200L}} & \multicolumn{2}{|c|}{\textbf{100d, Ftz, 800L}} & \multicolumn{2}{|c|}{\textbf{200d, Ftz, 800L}} \\
\hline
\textbf{Language} & \textbf{Prec.} & \textbf{Rec.} & \textbf{Prec.} & \textbf{Rec.} & \textbf{Prec.} & \textbf{Rec.} & \textbf{Prec.} & \textbf{Rec.} & \textbf{Prec.} & \textbf{Rec.} \\
\hline
\textit{W. Average} & \textit{0.99} & \textit{0.99} & \textit{0.99} & \textit{0.98} & \textit{0.99} & \textit{0.98} & \textit{0.95} & \textit{0.95} & \textit{0.95} & \textit{0.95} \\
\hline
Acehnese & 1.00 & 0.99 & 0.99 & 0.99 & 0.99 & 0.99 & 0.97 & 1.00 & 0.99 & 0.99 \\
Buginese & 1.00 & 0.98 & 0.99 & 0.98 & 0.99 & 0.98 & 0.99 & 0.97 & 0.98 & 0.98 \\
Cebuano & 1.00 & 1.00 & 1.00 & 1.00 & 1.00 & 1.00 & 1.00 & 1.00 & 0.99 & 1.00 \\
Chamorro & 0.96 & 1.00 & 0.75 & 1.00 & 0.67 & 1.00 & 0.58 & 1.00 & 0.52 & 1.00 \\
Chuukese & 1.00 & 1.00 & 1.00 & 1.00 & 1.00 & 1.00 & 1.00 & 1.00 & 1.00 & 1.00 \\
Fijian & 1.00 & 1.00 & 1.00 & 1.00 & 1.00 & 1.00 & 1.00 & 1.00 & 1.00 & 1.00 \\
Gilbertese & 1.00 & 1.00 & 1.00 & 1.00 & 1.00 & 1.00 & 1.00 & 1.00 & 1.00 & 1.00 \\
Hawaiian & 1.00 & 1.00 & 1.00 & 0.99 & 1.00 & 1.00 & 0.99 & 1.00 & 1.00 & 1.00 \\
Hiligaynon & 0.99 & 1.00 & 1.00 & 1.00 & 1.00 & 1.00 & 1.00 & 0.99 & 1.00 & 0.90 \\
Hiri Motu & 1.00 & 1.00 & 1.00 & 1.00 & 1.00 & 1.00 & 0.97 & 1.00 & 0.99 & 1.00 \\
Ilocano & 1.00 & 0.99 & 1.00 & 0.99 & 1.00 & 0.99 & 0.99 & 0.99 & 1.00 & 0.99 \\
Javanese & 0.97 & 0.99 & 1.00 & 0.99 & 0.98 & 0.99 & 0.96 & 0.96 & 0.98 & 0.92 \\
Marshallese & 1.00 & 1.00 & 1.00 & 1.00 & 1.00 & 1.00 & 1.00 & 1.00 & 1.00 & 1.00 \\
Malagasy & 1.00 & 1.00 & 0.99 & 1.00 & 0.99 & 1.00 & 0.99 & 1.00 & 1.00 & 1.00 \\
Māori & 1.00 & 1.00 & 1.00 & 0.99 & 0.99 & 1.00 & 1.00 & 0.99 & 1.00 & 0.99 \\
Malay & 0.97 & 0.99 & 0.98 & 0.99 & 0.99 & 0.99 & 0.96 & 0.97 & 0.85 & 0.99 \\
Niuean & 1.00 & 1.00 & 0.99 & 1.00 & 0.99 & 1.00 & 1.00 & 0.99 & 1.00 & 1.00 \\
Pangasinan & 1.00 & 0.97 & 1.00 & 0.97 & 0.99 & 0.97 & 0.99 & 0.97 & 1.00 & 0.96 \\
Pohnpeian & 1.00 & 1.00 & 1.00 & 1.00 & 1.00 & 1.00 & 1.00 & 1.00 & 1.00 & 1.00 \\
C.I. Māori & 1.00 & 1.00 & 1.00 & 1.00 & 1.00 & 0.99 & 0.99 & 1.00 & 1.00 & 1.00 \\
Samoan & 1.00 & 0.99 & 1.00 & 0.99 & 1.00 & 0.99 & 1.00 & 0.98 & 1.00 & 0.98 \\
Sundanese & 1.00 & 0.96 & 0.99 & 0.97 & 0.99 & 0.97 & 0.95 & 0.97 & 0.98 & 0.96 \\
Tahitian & 1.00 & 1.00 & 1.00 & 1.00 & 1.00 & 1.00 & 1.00 & 1.00 & 1.00 & 1.00 \\
Tagalog & 0.99 & 0.99 & 0.99 & 0.99 & 0.99 & 0.99 & 0.78 & 0.88 & 0.77 & 0.97 \\
Tongan & 1.00 & 0.97 & 1.00 & 0.94 & 1.00 & 0.95 & 0.99 & 0.96 & 1.00 & 0.95 \\
Tuvaluan & 1.00 & 1.00 & 1.00 & 1.00 & 1.00 & 0.99 & 1.00 & 1.00 & 1.00 & 1.00 \\
Waray & 1.00 & 1.00 & 1.00 & 1.00 & 1.00 & 1.00 & 0.99 & 1.00 & 1.00 & 1.00 \\
Wallisian & 1.00 & 1.00 & 1.00 & 1.00 & 0.99 & 1.00 & 0.99 & 1.00 & 1.00 & 1.00 \\
Yapese & 1.00 & 1.00 & 1.00 & 1.00 & 1.00 & 1.00 & 1.00 & 1.00 & 1.00 & 1.00 \\
\hline
  \end{tabular}
  \caption{Break-Down of \textsc{lid} Performance After Model Compression}
  \label{tab9}
\end{table*}

The second algorithm in Table \ref{tab:7} takes a word-level approach: the algorithm iteratres over each word in the sample, as before. In this case, however, the prediction is based mainly on word contexts rather than character spans that ignore words. \textit{First}, a baseline prediction is made on the entire sample. \textit{Second}, a prediction is made about the current word in isolation. \textit{Third}, a trigram prediction is made that includes the previous word and the following word. The algorithm combines these two predictions, the word and its context. If the overall sample is predicted to be English, the prediction closest to English is taken. But, if the overall sample is predicted to not be English, preference is given to a non-English prediction.

These two algorithms are used to convert span-level language predictions (covering 15-characters) into word-level predictions. Given the focus of the first portion of the paper on span-level language identification, the goal here is to enable code-switching detection for Austronesian languages using the same set of models.

\subsection{Evaluating Code-Switching Detection}

We evaluate the two algorithms described above using Wikipedia corpora that represent 14 Austronesian languages. These Wikipedia corpora are written mainly in the language of interest, but English terms remain in nearly every document. We therefore use the two code-switching detection algorithms and a baseline algorithm to predict sequences of three or more English words and then evaluate the accuracy of these sequences.

For each language, we retrieve the first 50 English phrases identified by each algorithm. We then use the short-span identification model to evaluate whether those spans as a whole are English (as they were predicted to be). The results are shown in Table \ref{tab:8}. The column \textit{Alg. 1} refers to the span-based algorithm in Table \ref{tab:7}. The column \textit{Alg. 2} refers to the word-based algorithm. And the column \textit{Baseline} refers to a simple model which directly queries the short-span model with each individual word (thus not taking any further context into consideration).

The word-based context algorithm has the highest accuracy, followed by the word-by-word baseline, with the character-span algorithm performing poorly overall. Three representative examples for the \textsc{eng-mri} model are shown below. For the character-based algorithm, words belonging to \textsc{mri} can be included in the predicted English span if the English words are rather long. For the word-based algorithm, almost all predicted English phrases are fully English, as shown in the example. Finally, the baseline -- which simply queries each word in isolation -- always identifies words like \textit{i} and \textit{he}, which could belong to either language, as belonging to English (which has more training data). Thus, sequences like the example below, when viewed as words in isolation, are identified incorrectly as English phrases.

~

\hspace{1cm}(Algorithm 1) \textit{tuatahi chant from the}
~

\hspace{1cm}(Algorithm 2) \textit{and are used to generate}
~

\hspace{1cm}(Baseline) \textit{i kōiwi hua he}
~

~

The evaluation in this section has shown that the underlying language identification models used here can also be adapted for short-span identification in a way that supports detection of code-switching in Austronesian languages. This is an important practical tool for supporting corpus creation because it allows us to control the proportion of a document that is in English. These models are available here: \href{https://github.com/jonathandunn/pacific_CodeSwitch}{https://github.com/jonathandunn/pacific\_CodeSwitch}

\section{Model Size and Model Stability}

Using fastText models leads to two practical challenges: first, the size of these models can become quite large \cite{joulin2016fasttext}; second, models based on skip-gram embeddings are subject to instability \cite{dunn-lrec}. A number of techniques for reducing model size are available, but each such technique has a chance of also reducing the performance of the model. This section provides an analysis of model performance under different compression strategies as well as the stability of the performance of the resulting models.

The results for compressed models are shown in Table \ref{tab9} for both 200-language and 800-language inventories. The leftmost column represents the original model (approximately 28 gb); the remaining columns evaluate lower dimensional models (100d, 200d) with 200-languages (200L) and 800-languages (800L). Each of these small models uses a minimum count threshold of 5 as well as having undergone quantization (represented here as \textit{Ftz}). These smaller models range from 600mb (200 languages with 100 dimensions) to 1.2gb (800 languages with 200 dimensions).

With 200 languages in the model, there is a slight decline in recall (from 0.99 in the original to 0.98 in both compressed models). Within Austronesian languages, this impact only is relevant for Tongan, which declines from 0.97 to 0.94 in recall. With 800 languages, there is a decline after compression to 0.95 precision and recall, compared with 0.96 in the original model in Table \ref{tab:4}. This slight decline in performance is necessary to have a model size which is small enough for practical use. The final models are available here: \href{https://jdunn.name/corpora/}{https://jdunn.name/corpora/}.

Given previous work showing the instability of the skip-gram embeddings that fastText uses, we might think that the performance of the language identification models here would also be unstable. To evaluate this we train five alternate versions of the 200 language compressed model and show the range of precision and recall values in Table \ref{tab:9} for each of the Austronesian languages. This experiment shows that two languages have a rather wide range of performance: Chamorro (which ranges from 0.60 to 0.76 precision), Sundanese (which ranges from 0.88 to 0.99 precision), and Hiligaynon (which ranges from 0.85 to 1.00 recall). Thus, the same instability which has an influence on unsupervised embeddings has an influence on supervised embeddings. In this case, we are able to select the model which has the best performance across all Austronesian languages.

\begin{table}[h]
\centering
\begin{tabular}{|l|cc|cc|}
\hline
\textbf{Language} &  \multicolumn{2}{|c|}{\textbf{Precision}} &  \multicolumn{2}{|c|}{\textbf{Recall}} \\
~ & \textit{Min} & \textit{Max} & \textit{Min} & \textit{Max} \\
\hline
Acehnese & 0.97 & 1.00 & 0.99 & 0.99 \\
Buginese & 0.99 & 1.00 & 0.98 & 0.98 \\
Cebuano & 0.99 & 1.00 & 1.00 & 1.00 \\
Chamorro & 0.60 & 0.76 & 1.00 & 1.00 \\
Chuukese & 1.00 & 1.00 & 1.00 & 1.00 \\
Fijian & 1.00 & 1.00 & 1.00 & 1.00 \\
Gilbertese & 1.00 & 1.00 & 1.00 & 1.00 \\
Hawaiian & 1.00 & 1.00 & 1.00 & 1.00 \\
Hiligaynon & 1.00 & 1.00 & 0.85 & 1.00 \\
Hiri Motu & 1.00 & 1.00 & 1.00 & 1.00 \\
Ilocano & 1.00 & 1.00 & 0.99 & 0.99 \\
Javanese & 0.97 & 1.00 & 0.99 & 0.99 \\
Marshallese & 1.00 & 1.00 & 1.00 & 1.00 \\
Malagasy & 0.99 & 1.00 & 1.00 & 1.00 \\
Māori & 0.99 & 1.00 & 0.97 & 1.00 \\
Malay & 0.96 & 0.99 & 0.99 & 0.99 \\
Niuean & 0.99 & 1.00 & 1.00 & 1.00 \\
Pangasinan & 0.99 & 1.00 & 0.96 & 0.97 \\
Pohnpeian & 1.00 & 1.00 & 1.00 & 1.00 \\
C. I. Māori & 0.97 & 1.00 & 1.00 & 1.00 \\
Samoan & 1.00 & 1.00 & 0.99 & 0.99 \\
Sundanese & 0.88 & 0.99 & 0.96 & 0.97 \\
Tahitian & 1.00 & 1.00 & 1.00 & 1.00 \\
Tagalog & 0.98 & 0.99 & 0.99 & 0.99 \\
Tongan & 1.00 & 1.00 & 0.94 & 0.96 \\
Tuvaluan & 1.00 & 1.00 & 0.98 & 1.00 \\
Waray & 1.00 & 1.00 & 1.00 & 1.00 \\
Wallisian & 1.00 & 1.00 & 1.00 & 1.00 \\
Yapese & 1.00 & 1.00 & 1.00 & 1.00 \\

\hline
  \end{tabular}
  \caption{Stability of Compressed Models (5x)}
  \label{tab:9}
\end{table}

\section{Conclusions}

This paper has evaluated and made available language identification models for previously unavailable Austronesian languages, both for identifying the language of documents as well as for identifying code-switching within documents. We have shown that the reduction of performance is minimal as the inventory of languages is systematically increased from 200 to 800. This work represents a significant advance in the availability of \textsc{nlp} resources for Austronesian languages.

% \nocite{*}
\section{Bibliographical References}\label{reference}

%\label{main:ref}

\bibliographystyle{lrec2022-bib}
\bibliography{lrec2022-example}

\begin{thebibliography}{}

\bibitem[\protect\citename{Agi{\'{c}} and
  Vuli{\'{c}}}2019]{agic-vulic-2019-jw300}
Agi{\'{c}}, {\v{Z}}. and Vuli{\'{c}}, I.
\newblock (2019).
\newblock {JW300: A Wide-Coverage Parallel Corpus for Low-Resource Languages}.
\newblock In {\em Proceedings of the Annual Meeting of the Association for
  Computational Linguistics}, pages 3204--3210. Association for Computational
  Linguistics, jul.

\bibitem[\protect\citename{Baroni \bgroup et al.\egroup }2009]{bbfz09}
Baroni, M., Bernardini, S., Ferraresi, A., and Zanchetta, E.
\newblock (2009).
\newblock {The WaCky Wide Web: A Collection of Very Large Linguistically
  Processed Web-crawled Corpora}.
\newblock {\em Language Resources and Evaluation}, 43(3):209--226.

\bibitem[\protect\citename{Benko}2014]{b14}
Benko, V.
\newblock (2014).
\newblock {Aranea Yet Another Family of (Comparable) Web Corpora}.
\newblock In {\em Proceedings of 17th International Conference Text, Speech and
  Dialogue.}, pages 257--264. Springer.

\bibitem[\protect\citename{Boyce}2006]{Boyce2006}
Boyce, M.
\newblock (2006).
\newblock {\em {A corpus of Modern Spoken Māori.}}
\newblock {Ph.D.} thesis, Victoria University of Wellington.

\bibitem[\protect\citename{Brown}2014]{Brown2014}
Brown, R.
\newblock (2014).
\newblock {Non-linear mapping for improved identification of 1300+ languages}.
\newblock In {\em Proceedings of the Conference on Empirical Methods in Natural
  Language Processing}, pages 627--632.

\bibitem[\protect\citename{Burdick \bgroup et al.\egroup
  }2021]{burdick-etal-2021-analyzing}
Burdick, L., Kummerfeld, J.~K., and Mihalcea, R.
\newblock (2021).
\newblock Analyzing the surprising variability in word embedding stability
  across languages.
\newblock In {\em Proceedings of the 2021 Conference on Empirical Methods in
  Natural Language Processing}, pages 5891--5901, Online and Punta Cana,
  Dominican Republic, November. Association for Computational Linguistics.

\bibitem[\protect\citename{Chakravarthi \bgroup et al.\egroup
  }2021]{chakravarthi-etal-2021-findings-vardial}
Chakravarthi, B.~R., Mihaela, G., Ionescu, R.~T., Jauhiainen, H., Jauhiainen,
  T., Lind{\'{e}}n, K., Ljube{\v{s}}i{\'{c}}, N., Partanen, N., Priyadharshini,
  R., Purschke, C., Rajagopal, E., Scherrer, Y., and Zampieri, M.
\newblock (2021).
\newblock {Findings of the VarDial Evaluation Campaign 2021}.
\newblock In {\em Proceedings of the Eighth Workshop on NLP for Similar
  Languages, Varieties and Dialects}, pages 1--11, Kiyv, Ukraine, apr.
  Association for Computational Linguistics.

\bibitem[\protect\citename{Dunn and Adams}2020]{dunn-adams-2020-geographically}
Dunn, J. and Adams, B.
\newblock (2020).
\newblock Geographically-balanced {G}igaword corpora for 50 language varieties.
\newblock In {\em Proceedings of the Language Resources and Evaluation
  Conference}, pages 2528--2536. European Language Resources Association, May.

\bibitem[\protect\citename{Dunn \bgroup et al.\egroup }2022]{dunn-lrec}
Dunn, J., Li, H., and Sastre, D.
\newblock (2022).
\newblock Predicting embedding reliability in low-resource settings using
  corpus similarity measures.
\newblock In {\em Proceedings of the 13th International Conference on Language
  Resources and Evaluation}. European Language Resources Association.

\bibitem[\protect\citename{Dunn}2020]{Dunn2020}
Dunn, J.
\newblock (2020).
\newblock {Mapping languages: the Corpus of Global Language Use}.
\newblock {\em Language Resources and Evaluation}, 54:999--1018.

\bibitem[\protect\citename{Goldhahn \bgroup et al.\egroup }2012]{geq12}
Goldhahn, D., Eckart, T., and Quasthoff, U.
\newblock (2012).
\newblock {Building Large Monolingual Dictionaries at the Leipzig Corpora
  Collection From 100 to 200 Languages}.
\newblock In {\em Proceedings of the Eighth Conference on Language Resources
  and Evaluation}, pages 759--765. European Language Resources Association.

\bibitem[\protect\citename{Jaech \bgroup et al.\egroup
  }2016]{jaech-etal-2016-hierarchical}
Jaech, A., Mulcaire, G., Hathi, S., Ostendorf, M., and Smith, N.
\newblock (2016).
\newblock {Hierarchical Character-Word Models for Language Identification}.
\newblock In {\em Proceedings of The Fourth International Workshop on Natural
  Language Processing for Social Media}, pages 84--93. Association for
  Computational Linguistics, nov.

\bibitem[\protect\citename{Jauhiainen \bgroup et al.\egroup
  }2017a]{jauhiainen-etal-2017-evaluating}
Jauhiainen, T., Lind{\'{e}}n, K., and Jauhiainen, H.
\newblock (2017a).
\newblock {Evaluating HeLI with Non-Linear Mappings}.
\newblock In {\em Proceedings of the Workshop on NLP for Similar Languages,
  Varieties and Dialects}, pages 102--108. Association for Computational
  Linguistics, apr.

\bibitem[\protect\citename{Jauhiainen \bgroup et al.\egroup
  }2017b]{jauhiainen-etal-2017-evaluation}
Jauhiainen, T., Lind{\'{e}}n, K., and Jauhiainen, H.
\newblock (2017b).
\newblock {Evaluation of language identification methods using 285 languages}.
\newblock In {\em Proceedings of the Nordic Conference on Computational
  Linguistics}, pages 183--191, Gothenburg, Sweden, may. Association for
  Computational Linguistics.

\bibitem[\protect\citename{Jauhiainen \bgroup et al.\egroup
  }2019]{Jauhiainen2019}
Jauhiainen, T., Lui, M., Zampieri, M., Baldwin, T., and Lind{\'{e}}n, K.
\newblock (2019).
\newblock {Automatic Language Identification in Texts: A Survey}.
\newblock {\em Journal of Artificial Intelligence Research}, 65.

\bibitem[\protect\citename{Joulin \bgroup et al.\egroup
  }2016]{joulin2016fasttext}
Joulin, A., Grave, E., Bojanowski, P., Douze, M., J{\'e}gou, H., and Mikolov,
  T.
\newblock (2016).
\newblock Fasttext.zip: Compressing text classification models.
\newblock {\em arXiv preprint arXiv:1612.03651}.

\bibitem[\protect\citename{Joulin \bgroup et al.\egroup }2017]{joulin2016bag}
Joulin, A., Grave, E., Bojanowski, P., and Mikolov, T.
\newblock (2017).
\newblock {Bag of Tricks for Efficient Text Classification}.
\newblock {\em Proceedings of the Conference of the European Chapter of the
  Association for Computational Linguistics}, pages 427--431.

\bibitem[\protect\citename{Kocmi and Bojar}2017]{kocmi-bojar-2017-lanidenn}
Kocmi, T. and Bojar, O.
\newblock (2017).
\newblock {LanideNN: Multilingual Language Identification on Character Window}.
\newblock In {\em Proceedings of the Conference of the European Chapter of the
  Association for Computational Linguistics}, pages 927--936. Association for
  Computational Linguistics, apr.

\bibitem[\protect\citename{Lison and
  Tiedemann}2016]{lison-tiedemann-2016-opensubtitles2016}
Lison, P. and Tiedemann, J.
\newblock (2016).
\newblock {OpenSubtitles2016: Extracting Large Parallel Corpora from Movie and
  TV Subtitles}.
\newblock In {\em Proceedings of the International Conference on Language
  Resources and Evaluation}, pages 923--929. European Language Resources
  Association, may.

\bibitem[\protect\citename{Lui and Baldwin}2011]{lui-baldwin-2011-cross}
Lui, M. and Baldwin, T.
\newblock (2011).
\newblock {Cross-domain Feature Selection for Language Identification}.
\newblock In {\em Proceedings of 5th International Joint Conference on Natural
  Language Processing}, pages 553--561. Asian Federation of Natural Language
  Processing, nov.

\bibitem[\protect\citename{Majl\u{i}s}2012]{Majlis2012}
Majl\u{i}s, M.
\newblock (2012).
\newblock {Yet Another Language Identifier}.
\newblock In {\em Proceedings of the EACL Student Research Workshop}, pages
  46--54. Association for Computational Linguistics.

\bibitem[\protect\citename{Malmasi and Dras}2017]{Malmasi2017}
Malmasi, S. and Dras, M.
\newblock (2017).
\newblock {Feature Hashing for Language and Dialect Identification}.
\newblock In {\em Proceedings of the Annual Meeting of the Association for
  Computational Linguistics}, pages 399--403.

\bibitem[\protect\citename{Molina \bgroup et al.\egroup
  }2016]{molina-etal-2016-overview}
Molina, G., AlGhamdi, F., Ghoneim, M., Hawwari, A., Rey-Villamizar, N., Diab,
  M., and Solorio, T.
\newblock (2016).
\newblock {Overview for the Second Shared Task on Language Identification in
  Code-Switched Data}.
\newblock In {\em Proceedings of the Second Workshop on Computational
  Approaches to Code Switching}, pages 40--49, Austin, Texas, nov. Association
  for Computational Linguistics.

\bibitem[\protect\citename{Solorio \bgroup et al.\egroup
  }2014]{solorio-etal-2014-overview}
Solorio, T., Blair, E., Maharjan, S., Bethard, S., Diab, M., Ghoneim, M.,
  Hawwari, A., AlGhamdi, F., Hirschberg, J., Chang, A., and Fung, P.
\newblock (2014).
\newblock {Overview for the First Shared Task on Language Identification in
  Code-Switched Data}.
\newblock In {\em Proceedings of the First Workshop on Computational Approaches
  to Code Switching}, pages 62--72, Doha, Qatar, oct. Association for
  Computational Linguistics.

\bibitem[\protect\citename{Tiedemann}2012]{Tiedemann2012}
Tiedemann, J.
\newblock (2012).
\newblock {Parallel Data, Tools and Interfaces in OPUS}.
\newblock In {\em Proceedings of the International Conference on Language
  Resources and Evaluation}, page 2214–2218. European Language Resources
  Association.

\end{thebibliography}

\section{Language Resource References}
\label{lr:ref}
\bibliographystylelanguageresource{lrec2022-bib}
\bibliographylanguageresource{languageresource}

\hangindent=1cm Dunn, J. (2022). \textit{Pacific Code-Switch: Python Package}. \href{https://github.com/jonathandunn/pacific_codeswitch}{https://github.com/jonathandunn/pacific\_CodeSwitch}.

~

\hangindent=1cm Dunn, J. (2022). \textit{Pacific Language  Identification Models}. \href{https://www.jdunn.name/corpora}{https://www.jdunn.name/corpora}.

\end{document}